\documentclass{article} % For LaTeX2e
\usepackage{iclr2026_conference}
\usepackage{times}

\usepackage{amsmath,amsfonts,bm}

\def\eqref#1{equation~\ref{#1}}
\def\1{\bm{1}}

\def\vc{{\bm{c}}}

\def\vk{{\bm{k}}}

\def\vq{{\bm{q}}}

\def\vv{{\bm{v}}}

\def\vx{{\bm{x}}}
\def\vy{{\bm{y}}}
\def\vz{{\bm{z}}}

\def\mA{{\bm{A}}}

\DeclareMathAlphabet{\mathsfit}{\encodingdefault}{\sfdefault}{m}{sl}
\SetMathAlphabet{\mathsfit}{bold}{\encodingdefault}{\sfdefault}{bx}{n}

\newcommand{\R}{\mathbb{R}}

\iclrfinalcopy
\usepackage{hyperref}
\usepackage{url}
\usepackage{float}
\usepackage{booktabs}
\usepackage{multirow}
\usepackage{graphicx}
\usepackage{bm}
\usepackage{tabularx}
\usepackage{amsmath}
\usepackage{adjustbox}
\usepackage{wrapfig}
\usepackage{caption}
\usepackage{subcaption}
\usepackage{setspace}

\title{MoGIC: Boosting Motion Generation via Intention Understanding and Visual Context}

\author{Junyu Shi, Yong Sun, Zhiyuan Zhang, Lijiang Liu, Zhengjie Zhang, Yuxin He, Qiang Nie\thanks{Corresponding author} \\
Robotics and Autonomous Systems \\
The Hong Kong University of Science and Technology (Guangzhou) \\
\texttt{jshi890@connect.hkust-gz.edu.cn} \\
}

\begin{document}

\maketitle

\begin{abstract}
% Human motion generation is central to animation, robotics, and embodied intelligence, yet existing text-driven methods often neglect human intention, producing motions that match semantics but lack coherent motivation. The absence of visual input further limits precision and personalization, as language alone is inherently ambiguous and insufficient to capture fine-grained spatiotemporal details. We propose MoGIC, a unified framework that integrates intention prediction with motion synthesis, enhanced by visual priors. MoGIC jointly optimizes five tasks across multimodal motion generation and intention prediction. This equips MoGIC with multimodal understanding and versatile generative capability. To support this paradigm, we curate and automatically annotate Mo440H, a large-scale benchmark constructed from 22 high-quality motion datasets. Experiments show that after fine-tuning on specific tasks, MoGIC significantly improves text-conditioned motion generation over previous works, reducing FID by 38.6\% on HumanML3D and 34.6\% on Mo440H. It also outperforms LLM-based methods in motion captioning with a lightweight text head. Beyond accuracy, MoGIC enables intention prediction and vision-conditioned generation, pushing forward personalized motion synthesis and intention understanding. The code will be available.

Existing text-driven motion generation methods often treat synthesis as a bidirectional mapping between language and motion, but remain limited in capturing the causal logic of action execution and the human intentions that drive behavior. The absence of visual grounding further restricts precision and personalization, as language alone cannot specify fine-grained spatiotemporal details. We propose MoGIC, a unified framework that integrates intention modeling and visual priors into multimodal motion synthesis. By jointly optimizing multimodal-conditioned motion generation and intention prediction, MoGIC uncovers latent human goals, leverages visual priors to enhance generation, and exhibits versatile multimodal generative capability. We further introduce a mixture-of-attention mechanism with adaptive scope to enable effective local alignment between conditional tokens and motion subsequences. To support this paradigm, we curate Mo440H, a 440-hour benchmark from 21 high-quality motion datasets. Experiments show that after finetuning, MoGIC reduces FID by 38.6\% on HumanML3D and 34.6\% on Mo440H, surpasses LLM-based methods in motion captioning with a lightweight text head, and further enables intention prediction and vision-conditioned generation, advancing controllable motion synthesis and intention understanding. The code is available at \url{https://github.com/JunyuShi02/MoGIC}

\end{abstract}
\section{Introduction}
Human motion generation has emerged as a central research direction in artificial intelligence, with broad applications in animation, virtual reality, and embodied intelligence. Recent advances in generative modeling have enabled text-conditioned methods to synthesize realistic human motions from natural language descriptions. However, existing approaches often overlook the role of human intention and rely on limited modalities, leading to shortcomings in controllability, and generalization.

Most text-driven motion generation methods~\cite{guo2024momask, meng2025mardm} treat synthesis as a one-way mapping from text to motion, without explicitly modeling human intention, which often constitutes the underlying logic of action execution. Recent GPT-based approaches~\cite{jiang2023motiongpt, wu2025mg, luo2024m, Jiang2025motionchain} extend this paradigm by modeling discretized motion as a second language, thereby enabling bidirectional generation between motion and text. In this context, text generation can be regarded as a means of interpreting human intention. However, such bidirectional modeling reduces the motion–language association to a cross-modal mapping and remains limited in capturing the causal logic that governs how motions unfold. Consequently, despite leveraging large-scale linguistic priors and demonstrating strong generative capacity, these models remain less effective than specialized motion generation frameworks. Intuitively, a model should not only understand how actions are performed. At a higher semantic level, it should also predict the latent human intention underlying unobserved motions. This enables the model to uncover the internal causal structure that drives the occurrence of motion.

Another critical limitation lies in the absence of visual modality, which leads to motions that lack precision and personalization. Language alone is inherently ambiguous and cannot fully specify fine-grained spatiotemporal details; even when such details are explicitly described, their symbolic form does not readily translate into precise control signals. For instance, the instruction “a person walks forward, turns left, then picks up an apple” leaves unspecified the walking speed, turning angle, and spatial location of the apple, forcing the model to sample from a highly uncertain distribution and often producing unsatisfactory results. By contrast, visual inputs naturally provide joint trajectories and environmental context, such as human–scene or human–object interactions. Without such visual grounding, models fail to exploit these priors, resulting in less accurate motions and restricting their applicability to text-only conditioned generation.

To overcome these limitations, we introduce MoGIC, a unified framework that integrates multi-conditional control with explicit intent modeling. Building on the observation that motions are not only shaped by surface semantics but also by underlying goals, MoGIC leverages fine-grained textual descriptions of incomplete motions as human intentions. The model is equipped with disentangled generation heads: the Intention Prediction Head outputs discrete descriptions that specify what to do (goals), while the Motion Generation Head produces continuous trajectories that express how to do (execution). Through this disentangled generation process, MoGIC jointly optimizes intention and motion while avoiding semantic confusion caused by their heterogeneous representations. This design yields substantial improvements in motion quality, achieving a 35.2\% reduction in FID compared to training without intention prediction on MoGIC.

To further address the ambiguity of purely textual conditioning, we incorporate visual modality as an additional condition. Unlike pose estimation methods that reconstruct dense trajectories, we use low-frame-rate image sequences as weak but informative priors. These visual conditions introduce auxiliary perceptual priors into the training process, thereby enhancing the quality of representation learning and yielding consistent improvements across downstream tasks. Furthermore, vision provides conditions that alleviate the inherent ambiguity of language, enabling the generation of motions that are more precise and controllable. Moreover, the integration of vision unlocks the model’s capacity to handle a wider range of tasks beyond text-only conditioning, such as vision-conditioned motion completion or generating diverse motions from sparse video frame.

Our video inputs (sampled at 1 fps) are not temporally aligned with motion sequences (30 fps). Instead of enforcing rigid frame-to-frame correspondence, we leverage them as priors for trajectory and scene context. Similarly, text only partially aligns with motion, as some phrases correspond to specific fragments. To handle such partial correspondences, we introduce a mixture-of-attention mechanism with adaptive scope, enabling motion tokens to interact with the most relevant conditional tokens across granularities. This strengthens local–global alignment and mitigates confusion from temporal mismatches or ambiguous conditions.

We adopt a two-stage training strategy. In the first stage, MoGIC learns cross-modal consistency by jointly optimizing multimodal-conditioned motion generation loss and intention prediction loss, enabling robust generalization across different input types. In the second stage, we finetune on task-specific objectives such as language-to-motion generation, effectively transferring visual and intentional priors to downstream tasks. 

To enable tri-modal learning, we curate and automatically annotate 21 high-quality motion datasets into a large-scale benchmark, Mo440H, which spans 440 hours of single-person motions, human–human interactions, and human–object interactions. Extensive experiments on both HumanML3D and Mo440H show that MoGIC achieves substantial gains over SOTA methods, reducing FID by 38.6\% and 34.6\% respectively. Moreover, fine-tuning on the Motion Caption task demonstrates that MoGIC, with its lightweight text generation head, surpasses LLM-based baselines despite using fewer parameters. Furthermore, our experiments demonstrate that incorporating the visual modality not only enhances controllability and overall generation quality, but also unlocks new capabilities such as image-to-motion synthesis and vision-conditioned motion completion.

\section{Related Work}
\paragraph{Motion Generation} Recent work on text-conditioned motion generation has primarily relied on probabilistic generative models, including GANs~\cite{harvey2020robust, ghosh2021synthesis}, VAEs~\cite{petrovich2022temos}, and diffusion methods~\cite{du2023avatars, chen2023executing, shafir2024human, tevet2023human, zhang2023finemogen, zhang2023remodiffuse, zhang2024motiondiffuse, xie2024omnicontrol, zhou2025emdm, meng2025mardm}, which generate realistic motion by sampling and refining noise. In parallel, discrete-token approaches~\cite{du2023avatars, zhong2023attt2m, zhang2023generating, pinyoanuntapong2024bamm, shi2025genm, wang2025motiondreamer, jeong2025hgm} use vector-quantized autoencoders to construct a motion vocabulary, with transformers modeling token sequences either autoregressively or through masked denoising. 

More recent studies have begun to explore the use of large language models to capture the rich temporal and semantic structure of human motion~\cite{jiang2023motiongpt, luo2024m, liang2024omg, wang2024motiongpt, wu2025motionagent}. By treating motion as a discretized foreign language, these methods learn bidirectional mappings between text-to-motion (T2M) and motion-to-text (M2T). However, such mappings remain at the level of superficial pattern matching, and the potential of bidirectional alignment to improve motion generation has not been fully explored. In this paper, we advance motion generation by enabling the model to infer the underlying causes of motion, namely human intentions.

More modalities have also been explored as conditions. For instance, \cite{chen2025language} unify audio and text representations of 3D human motions using LLMs, while MotionAnything~\cite{zhang2025motion} combines text and music to generate more controllable dance motions. Such modalities provide complementary information, yet the crucial role of visual inputs in motion generation remains underexplored. In this paper, we incorporate visual modality to reduce the ambiguity of text-only descriptions and extend the model’s applicability to a wider range of downstream tasks.

\paragraph{Motion Dataset} In recent years, numerous motion datasets have emerged to capture diverse human activities. AMASS~\cite{mahmood2019amass} provides high-quality 3D mocap data, further extended by BABEL~\cite{punnakkal2021babel} with segment-level categorical labels and HumanML3D~\cite{guo2022humanml3d} with sequence-level textual descriptions. Beyond these, many datasets~\cite{mehta2017monocular, fieraru2021aifit, cai2022humman, xiong2024mvhumannet, liu2022beat, tripathi20233d} capture a broad spectrum of movements ranging from daily actions to gestures and yoga. EgoBody~\cite{zhang2022egobody} and InterGen~\cite{liang2024intergen} focus on two-person interactions, while others~\cite{lv2025himo, zhao2024imhoi, li2023object, jiang2023full, taheri2020grab} emphasize human–object interactions. Collaborative and scene-aware scenarios are also addressed by datasets such as CORE4D~\cite{liu2024core4d}, Humanise~\cite{wang2022humanise}, PROX~\cite{hassan2019resolving}, and HOI-M3~\cite{zhang2024hoi}.

Recent works~\cite{lin2023motionx, lu2025scamo, fan2025go} extract motions from online videos via pose estimation. However, this approach results in lower fidelity and exhibits content bias. For instance, MotionMillion~\cite{fan2025go} contains more than 70\% sports-related activities such as martial arts, fitness, and dance. In this paper, we integrate and re-annotate over twenty high-quality motion datasets, achieving large-scale coverage while avoiding the quality issues and content bias of video-derived data.
\begin{figure}[t]
\centering 
\includegraphics[width=1\columnwidth]{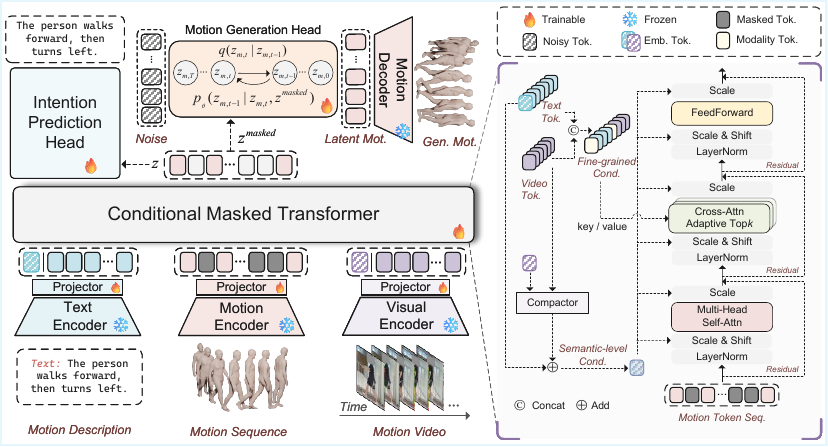} 
\vspace{-0.6cm}
\caption{Overview of MoGIC. The framework consists of modality-specific encoders, a Conditional Masked Transformer (CMT), a Motion Generation Head (MGH), and an Intention Prediction Head (IPH). Language, vision, and motion inputs are first processed by their respective encoders to produce latent tokens. Motion tokens are randomly masked and passed through the CMT, where semantic-level and fine-grained conditions modulate the motion token in series. The resulting conditional tokens $z$ are used in two branches: (i) the masked motion tokens are reconstructed via the MGH, which denoises them into clean motion latent tokens and decodes them into motion sequences; (ii) $z$ serves as key and query signals for the IPH to predict the underlying intention.}
% \vspace{-5mm}
\label{fig:pipeline}
\vspace{-0.5cm}
\end{figure}

\section{Model Architecture}
We propose MoGIC, a unified framework for human motion and intention generation conditioned on arbitrary multimodal inputs, including language, vision, and partially visible motion sequences. As shown in Figure~\ref{fig:pipeline}, each modality is first projected into the latent space via modality-specific encoders, with random masking applied to motion tokens for generative masked modeling. A Conditional Masked Transformer (CMT) then integrates the projected conditioning signals at both semantic-level and fine-grained conditions to modulate the masked motion tokens. The resulting motion tokens serve as a unified representation that generates both high-level intentions and complete motion latent sequences, which are subsequently reconstructed into the original motion domain through a motion decoder.

\subsection{Modality-Specific Encoder}

\paragraph{Motion/Language/Vision Encoder} 
Given a motion sequence $\vx_m \in \R^{l_m \times c_m}$ with $l_m$ frames and $c_m$ feature dimensions, we adopt a temporal convolutional auto-encoder to project motion into a compact latent space $\vz_m = f_m(\vx_m) \in \R^{l_m^{'} \times d_m}$, where $l_m^{'} \leq l_m$ and $d_m$ is the motion embedding dimension. A symmetric decoder reconstructs the motion as $\hat{\vx}_m = g_m(\vz_m)$, and the auto-encoder is trained with a smooth L1 reconstruction loss $\mathcal{L}_{\text{rec}} = \text{SmoothL1}(\hat{\vx}_m, \vx_m)$. For text, a frozen CLIP encoder outputs token-level embeddings $\vz_t = f_t(\vx_t) \in \R^{l_t \times d_t}$, where the \texttt{[CLS]} token $\vz_t^g$ captures global semantics. For vision, video frames sampled at 1 fps are encoded as $\vz_v^{p} = f_v(\vx_v^p) \in \R^{d_v}$, and aggregated into a global representation by attention layer $\vz_v^g = \operatorname{attn}(\vq_v, \vz_v, \vz_v)$ with a learnable query vector $\vq_v$.

\subsection{Conditional Masked Transformer}
The Conditional Masked Transformer integrates multimodal conditioning signals into motion tokens through two operations: (i) global semantic-level modulation, which injects fused text–vision context into the motion representation via adaptive normalization to ensure semantic consistency, and (ii) mixture-of-attention with adaptive scope, realized through the adaptive Top-$k$ cross-attention mechanism that dynamically aligns motion tokens with the most relevant text-vision snippets while adaptively determining the scope of attended context. At each layer of the CMT, motion tokens first pass through a self-attention module, then receive fine-grained conditions via the adaptive Top-$k$ cross-attention, and finally go through a feed-forward network to produce the output representation.

\paragraph{Semantic-level Modulation} 
We adopt adaptive LayerNorm modulation. The global multimodal context vector $\vc^{g} = \vz_t^{g} + \vz_v^{g}$ is mapped to modulation coefficients $(\alpha_c, \beta_c, \gamma_c)  = W_{\text{ada}}(\vc^{g}) \in \R^{d_m}$ via a lightweight MLP. Given a normalized motion token $\bar{\vz}_m = \operatorname{LN}(\vz_m)$, modulation and gated residual connection are applied as
\begin{equation}
    \vz_m \gets \vz_m + \gamma_c \odot h(\alpha_c \odot \bar{\vz}_m + \beta_c)
\end{equation}
where $h(\cdot)$ denotes the corresponding sub-layer transformation (self-attention, cross-attention, and feed-forward layer in CMT). This formulation ensures that global multimodal context consistently modulates motion representation while flexibly controlling residual pathways.

\paragraph{Mixture-of-Attention with Adaptive Scope}
To enable fine-grained dynamic alignment between motion tokens and multimodal conditioning signals, we employ the mixture-of-attention mechanism that operates on concatenated token-level condition embeddings $\vc^{tok} = [\vz_v; \vz_t] \in \R^{(p + l) \times d}$. When a modality is missing, its slot in $\vc^{tok}$ is replaced by a learnable embedding. Given query motion tokens $\vz_m \in \R^{l_m^{'} \times d}$, each expert computes queries, keys, and values as $\vq^{e} = W_q^{e} \vz_m$, $\vk^{e} = W_k^{e} \vc^{tok}$, and $\vv^{e} = W_v^{e} \vc^{tok}$, followed by attention to produce the score matrix $\mA^{e}$. To control the effective context scope, we sort $\mA^{e}$ in descending order per query and accumulate until a cumulative mass $\tau$ is reached:
\begin{equation}
k_{\text{dyn}}^{e} = \min\Big(\max\big( \arg\min_k { \textstyle\sum_{j=1}^k \mA_{(j)}^{e} \geq \tau }, k_{\min}^{e}\big), k_{\max}^{e}\Big)
\end{equation}
where $\mA_{(j)}^{e}$ denotes the $j^{th}$ largest weight. The final attention distribution is restricted to the top-$k_{\text{dyn}}^{e}$ entries. The final output is the sum of all expert contributions
\begin{equation}
\vz = \sum_{e=1}^{E} \tilde{\mA}^e \vv^e, \quad \tilde{\mA}_{i,j}^{e} =
\frac{\mA_{i,j}^{e} \cdot \mathbf{1}{ j \in \text{top-}k_{\text{dyn}}^{e}(i)}}
{\sum_{j' \in \text{Top-}k_{\text{dyn}}(i)} \mA_{i,j'}^{e}}
\end{equation}
where $\text{top-}k_{\text{dyn}}^{e}(i)$ denotes the set of indices of the attention weights for the $i^{th}$ motion token. This adaptive mixture-of-attention design ensures that motion tokens selectively attend to the most relative condition tokens, while maintaining flexibility to balance semantic consistency and fine-grained alignment across diverse contexts.

\subsection{Disentangle Generation Head}
Human intention and motion represent two fundamentally different semantic levels, with intention being linguistically oriented and motion encoding continuous dynamics. To capture this distinction, we adopt disentangled generation heads that separately model the two modalities
\paragraph{Intention Prediction Head (IPH)}
The Intention Prediction Head (IPH) captures human intention in explicit textual form. It employs a T5-style~\cite{raffel2020exploring} decoder that, conditioned on the embedding $\vz$ from the conditional masked transformer, generates the intention description in an autoregressive manner. Each decoder layer combines self-attention over the partially generated sequence with cross-attention conditioned on $\vz$
\paragraph{Motion Generation Head (MGH)}
Since motions are continuous, we employ a continuous-time interpolant model in the Motion Generation Head (MGH) following SiT~\cite{ma2024sit}, conditioned on the masked embedding $\vz^{mask}$. The interpolant at time $t \in [0,1]$ is defined as:
\begin{equation}
\begin{aligned}
\vz_{m,t} &= \alpha_t \vz_{m,0} + \sigma_t \boldsymbol{\epsilon}, 
\quad \boldsymbol{\epsilon} \sim \mathcal{N}(0, I) \\
\vv_\theta(\vz_{m,t}, t, \vz^{mask}) &\approx \dot{\alpha}_t \vz_{m,0} + \dot{\sigma}_t \boldsymbol{\epsilon}
\end{aligned}
\end{equation}
where $\vz_{m,0}$ is the latent motion of ground truth and $\vv_{\theta}$ predicts the velocity field under conditioning.
For sampling, given $\vz^{mask}$, we integrate the learned velocity field backward in time using either the probability flow ODE or reverse-time SDE:
\begin{equation}
\vz_{m,0} = \mathrm{Solver}(\vz_{m,T}, \vv_\theta, \vz^{mask}), \quad \vz_{m,T} \sim \mathcal{N}(0,I).
\end{equation}
The denoised latent representation $\vz_{m,0}$ is subsequently provided to the motion decoder to reconstruct the full motion sequence $\hat{\vx}_m$.

\section{Cross-Modal Generative Training}
We train MoGIC on five tasks: (1) language-to-motion (L2M), which generates motion from textual descriptions; (2) vision-language-to-motion (VL2M), which integrates visual and textual inputs to produce more controllable motion; (3) vision-to-motion (V2M), which synthesizes motion purely from visual sequences; (4) motion-to-motion (M2M), which reconstructs complete motion from partially observed sequences; and (5) intention prediction (IP), which infers high-level motivational factors behind motion. All tasks share a Conditional Masked Transformer with modality-specific conditioning. Motion sequences are encoded into latent tokens $\vz_m \in \R^{l_m' \times d_m}$, where a subset is randomly masked with learnable tokens for generative reconstruction, and for intention prediction, the latter 50\% of tokens are additionally truncated. The fused masked sequence and modalities yield a motion embedding $z$, which conditions both the IPH and MGH for intention prediction and motion generation. Training is driven by a joint loss combining a diffusion-based velocity matching objective for motion and an autoregressive cross-entropy for intention:
\begin{equation}
    \mathcal{L} =
    \lambda_{\text{motion}}\,
    \mathbb{E}_{t,\boldsymbol{\epsilon}}\Big[
    \big\|
    \vv_\theta(\vz_{m,t}, t, \vz) -
    (\dot{\alpha}_t \vz_{m,0} + \dot{\sigma}_t \boldsymbol{\epsilon})
    \big\|_2^2
    \Big]
    +
    \lambda_{\text{intent}}\,
    \mathbb{E}_{(\vy, \vz)}\Big[
    - \sum_{i=1}^{T} \log P(y_i|y_{<i}, \vz)
    \Big]
\end{equation}
This unified training framework enables the model to learn a shared latent space where motion and intention are jointly optimized. The decoupled intention prediction guides the model to capture the underlying motivational factors of motion, while mitigating the semantic entanglement between discrete text and continuous motion representations.

\section{Experiments}

\setlength{\tabcolsep}{2pt}
\begin{table}[t]
\captionsetup{skip=3pt}
\scriptsize
\centering
\caption{The quantitative results of L2M on HumanML3D. The best results are displayed in bold. Noting that the metric values of some methods are adopted from MARDM~\citep{meng2025mardm}.}
\begin{adjustbox}{width=\textwidth}
\begin{tabular}{lcccccc}
\toprule
\multirow{2}{*}{Methods} & 
\multicolumn{3}{c}{R Precision$\uparrow$} & 
\multirow{2}{*}{FID$\downarrow$} & 
\multirow{2}{1cm}{\centering Matching$\downarrow$} & 
\multirow{2}{1.5cm}{\centering CLIP-score $\uparrow$} \\
\cline{2-4}
& Top 1 & Top 2 & Top 3 \\
\midrule

T2M-GPT~\citep{zhang2023generating} & $0.470^{\pm .003}$ & $0.659^{\pm .002}$ & $0.758^{\pm .002}$ & $0.335^{\pm .003}$ & $3.505^{\pm .017}$ & $0.607^{\pm .005}$ \\
ReMoDiffuse~\citep{zhang2023remodiffuse} & $0.468^{\pm .003}$ & $0.653^{\pm .003}$ & $0.754^{\pm .005}$ & $0.883^{\pm .021}$ & $3.414^{\pm .020}$ & $0.621^{\pm .003}$ \\
MDM-50Step~\citep{tevet2023human} & $0.440^{\pm .007}$ & $0.636^{\pm .006}$ & $0.742^{\pm .004}$ & $0.518^{\pm .032}$ & $3.640^{\pm .028}$ & $0.578^{\pm .003}$ \\
MLD~\citep{chen2023executing} & $0.461^{\pm .004}$ & $0.651^{\pm .004}$ & $0.750^{\pm .003}$ & $0.431^{\pm .014}$ & $3.445^{\pm .019}$ & $0.610^{\pm .003}$ \\
MMM~\citep{pinyoanuntapong2024mmm} & $0.487^{\pm .003}$ & $0.683^{\pm .002}$ & $0.782^{\pm .001}$ & $0.132^{\pm .004}$ & $3.359^{\pm .009}$ & $0.635^{\pm .003}$ \\
MoMask~\citep{guo2024momask} & $0.469^{\pm .004}$ & $0.687^{\pm .003}$ & $0.786^{\pm .003}$ & $0.116^{\pm .006}$ & $3.353^{\pm .010}$ & $0.637^{\pm .003}$ \\
MotionDiffuse~\citep{zhang2024motiondiffuse} & $0.450^{\pm .006}$ & $0.641^{\pm .005}$ & $0.753^{\pm .005}$ & $0.778^{\pm .005}$ & $3.490^{\pm .023}$ & $0.606^{\pm .004}$ \\
MARDM-DDPM~\citep{meng2025mardm} & $0.492^{\pm .006}$ & $0.690^{\pm .005}$ & $0.790^{\pm .005}$ & $0.116^{\pm .004}$ & $3.349^{\pm .010}$ & $0.637^{\pm .005}$ \\
MARDM-SiT~\citep{meng2025mardm} & $0.500^{\pm .004}$ & $0.695^{\pm .003}$ & $0.795^{\pm .003}$ & $0.114^{\pm .007}$ & $3.270^{\pm .009}$ & $0.642^{\pm .002}$ \\
MotionAgent~\cite{wu2025motionagent} & $0.485^{\pm .003}$ & $0.680^{\pm .003}$ & $0.780^{\pm .002}$ & $0.202^{\pm .009}$ & $3.327^{\pm .009}$ & $0.634^{\pm .003}$ \\
\midrule
MoGIC (ours) \textit{w/o Int. loss}  & $0.533^{\pm 0.012}$ & $0.731^{\pm 0.010}$ & $0.826^{\pm 0.010}$ & $0.108^{\pm 0.023}$ & $3.078^{\pm 0.037}$ & $0.658^{\pm 0.001}$ \\
MoGIC (ours) & $\bm{0.545^{\pm 0.003}}$ & $\bm{0.741^{\pm 0.003}}$ & $\bm{0.835^{\pm 0.002}}$ & $\bm{0.070^{\pm 0.004}}$ & $\bm{2.999^{\pm 0.011}}$ & $\bm{0.669^{\pm 0.001}}$ \\

\bottomrule
\end{tabular}
\end{adjustbox}
\label{tab:t2m-h3d}
% \vspace{-0.3cm}
\end{table}

\setlength{\tabcolsep}{2pt}
\begin{table}[t]
\captionsetup{skip=3pt}
\scriptsize
\centering
\caption{The quantitative results of L2M on Mo440H-ML. The best results are displayed in bold. }
\begin{adjustbox}{width=\textwidth}
\begin{tabular}{lcccccc}
\toprule
\multirow{2}{*}{Methods} & 
\multicolumn{3}{c}{R Precision$\uparrow$} & 
\multirow{2}{*}{FID$\downarrow$} & 
\multirow{2}{1cm}{\centering Matching$\downarrow$} & 
\multirow{2}{1.5cm}{\centering Diversity$\uparrow$} \\
\cline{2-4}
& Top 1 & Top 2 & Top 3 \\
\midrule

MotionDiffuse~\citep{zhang2024motiondiffuse} & $0.550^{\pm .001}$ & $0.735^{\pm .001}$ & $0.801^{\pm .002}$ & $0.957^{\pm .010}$ & $2.990^{\pm .007}$ & $12.009^{\pm .104}$ \\
MMM~\citep{pinyoanuntapong2024mmm} & $0.601^{\pm .001}$ & $0.798^{\pm .001}$ & $0.887^{\pm .001}$ & $0.237^{\pm .004}$ & $2.420^{\pm .004}$ & $11.883^{\pm .089}$ \\
MoMask~\citep{guo2024momask} & $0.610^{\pm .001}$ & $0.801^{\pm .002}$ & $0.886^{\pm .001}$ & $0.205^{\pm .006}$ & $2.353^{\pm .003}$ & $11.963^{\pm .077}$ \\
MARDM-DDPM~\cite{meng2025mardm} & $0.573^{\pm .001}$ & $0.785^{\pm .002}$ & $0.885^{\pm .002}$ & $0.431^{\pm .004}$ & $2.166^{\pm .005}$ & $\bm{12.630^{\pm .079}}$ \\
MARDM-SiT~\cite{meng2025mardm} & $0.613^{\pm .001}$ & $0.820^{\pm .002}$ & $0.906^{\pm .001}$ & $0.231^{\pm .003}$ & $2.420^{\pm .005}$ & $12.112^{\pm .079}$ \\
MG-MotionLLM~\cite{wu2025mg} & $0.556^{\pm 0.002}$ & $0.737^{\pm 0.002}$ & $0.834^{\pm 0.002}$ & $0.624^{\pm 0.008}$ & $2.544^{\pm 0.006}$ & $12.252^{\pm 0.099}$ \\
\midrule

MoGIC (ours) \textit{only L2M loss} & $0.637^{\pm 0.001}$ & $0.836^{\pm 0.001}$ & $0.908^{\pm 0.002}$ & $0.201^{\pm 0.001}$ & $2.003^{\pm 0.007}$ & $12.392^{\pm 0.084}$ \\ 
MoGIC (ours) \textit{L2M + Int. loss} & $\bm{0.652^{\pm 0.001}}$ & $\bm{0.851^{\pm 0.001}}$ & $\bm{0.926^{\pm 0.001}}$ & $0.134^{\pm 0.001}$ & $\bm{1.889^{\pm 0.005}}$ & $12.434^{\pm 0.087}$ \\
MoGIC (ours) \textit{L2M + Caption loss} & $0.646^{\pm 0.001}$ & $0.845^{\pm 0.001}$ & $0.919^{\pm 0.001}$ & $0.198^{\pm 0.001}$ & $1.910^{\pm 0.005}$ & $12.623^{\pm 0.090}$ \\
MoGIC (ours) & $0.643^{\pm 0.001}$ & $0.844^{\pm 0.002}$ & $0.917^{\pm 0.002}$ & $0.185^{\pm 0.002}$ & $1.915^{\pm 0.004}$ & $12.516^{\pm 0.077}$ \\
MoGIC (ours) \textit{w/ L2M FT} & $0.651^{\pm 0.001}$ & $0.849^{\pm 0.001}$ & $0.924^{\pm 0.002}$ & $\bm{0.123^{\pm 0.001}}$ & $1.903^{\pm 0.006}$ & $12.511^{\pm 0.091}$ \\

\bottomrule
\end{tabular}
\end{adjustbox}
\label{tab:t2m-mydataset}
\vspace{-0.4cm}
\end{table}

\subsection{Integrated Motion Dataset}
\paragraph{Motion Dataset} We curated and processed 21 high-quality motion datasets covering diverse scenarios such as single-person activities, human–human interactions, and human–object interactions. All motions were standardized to a 22-joint format, resampled to 30 fps, and capped at 10 seconds. For datasets without textual annotations but with visual modalities, we used Qwen2.5-VL-Max~\cite{ali2025Qwen2.5-VL} to generate captions and manually filtered inadequate samples; for those lacking RGB videos, rendered mesh sequences were adopted instead, with all videos downsampled to 1 fps. The final collection, termed Mo440H, comprises about 440 hours of motion (about 50M frames), 210k textual descriptions, and 140k image sequences. Depending on available modalities, we further organize it into three subsets: Mo440H-All (the whole dataset, for auto-encoder training and cross-modal generative training), Mo440H-ML (motion–language pairs, for language-to-motion and motion-to-language), and Mo440H-MLV (motion–language–vision triplets, enabling visually conditioned tasks).

In addition, we evaluate on the HumanML3D~\cite{guo2022humanml3d} dataset, a widely used benchmark with about 14k motion sequences and 45k text annotations, following established protocols~\cite{meng2025mardm} for fair comparison with previous work.

\paragraph{Motion Representation} We adopt a compact motion representation by removing redundant features (e.g., 6D rotations), following~\cite{meng2025mardm}, to mitigate distribution mismatch and generation errors. The motion data is represented as $\vx_m^i = [\dot{r}^a, \dot{r}^{xz}, \dot{r}^h, j^p]$ at time step $i$, consisting of root angular velocity $\dot{r}^a$, root linear velocities $\dot{r}^{xz}$ in the XZ-plane, root height $\dot{r}^h$, and local joint positions $j^p \in \mathbb{R}^{3(N_j-1)}$, which jointly encode the essential kinematic information for motion.

\subsection{Experiment Settings and Evaluation Metrics}
\paragraph{Experiment Settings} All experiments are conducted on RTX4090 GPUs with a batch size of 64 using the Adam optimizer (lr=2e-4, 2000-step warm-up), training for 500 epochs on HumanML3D and 10M iterations on Hu440H ($\approx$40 GB GPU memory). The motion generation loss is optimized every epoch, while the intention prediction loss is updated every 4 epochs. The Conditional Masked Transformer (384 channels) uses 1 layer for HumanML3D and 2 layers for Hu440H dataset. Cross-attention employs two parallel modules ($k \in [1,6]$, threshold 0.8; and $k \in [0,\infty]$, threshold 1). The intention head is a 3-layer T5-style decoder, and the motion head is a diffusion model with a 10-layer MLP (1280 channels).

\begin{wraptable}{r}{0.55\textwidth}
    \vspace{-1cm}
    \captionsetup{skip=3pt}
    % \scriptsize
    \centering
    \caption{Comparisons of motion in-between tasks on Mo440H-ML. Each setting reports R-precision top 3 (R@3), FID, and Matching score (Match).}
    \resizebox{0.55\textwidth}{!}{
        \begin{tabular}{llcccccc}
        \toprule
        \multirow{3}{*}{\textbf{Task}} & \multirow{3}{*}{\textbf{Method}} & 
        \multicolumn{3}{c}{\textbf{w/o language}} & \multicolumn{3}{c}{\textbf{w/ language}} \\
        \cmidrule(lr){3-5} \cmidrule(lr){6-8}
        & & R@3$\uparrow$ & FID$\downarrow$ & Match$\downarrow$
          & R@3$\uparrow$ & FID$\downarrow$ & Match$\downarrow$ \\
        \midrule
        
        \multirow{2}{*}{pref.} 
            & MARDM      & $0.874$ & $0.286$ & $2.808$    & $0.912$ & $0.194$ & $1.972$ \\
            & MoGIC      & $0.892$ & $0.173$ & $2.172$    & $0.943$ & $0.128$ & $1.644$ \\
        \midrule
        
        \multirow{2}{*}{suff.} 
            & MARDM      & $0.894$ & $0.239$ & $2.334$    & $0.912$ & $0.188$ & $1.989$ \\
            & MoGIC      & $0.912$ & $0.140$ & $1.938$    & $0.941$ & $0.091$ & $1.647$ \\
        \midrule
        
        \multirow{2}{*}{inf.}  
            & MARDM      & $0.907$ & $0.211$ & $2.249$    & $0.913$ & $0.186$ & $1.984$ \\
            & MoGIC      & $0.926$ & $0.124$ & $1.789$    & $0.943$ & $0.113$ & $1.619$ \\
        \midrule
        
        \multirow{2}{*}{circ.} 
            & MARDM      & $0.896$ & $0.249$ & $2.358$    & $0.913$ & $0.175$ & $1.980$ \\
            & MoGIC      & $0.912$ & $0.147$ & $1.979$    & $0.943$ & $0.109$ & $1.639$ \\
        \bottomrule
        \end{tabular}
    }
    \label{tab:in-between-all}
    \vspace{0.2cm}
% \end{wraptable}

% \begin{wraptable}{r}{0.5\textwidth}
  % \vspace{-0.1cm}
  \caption{Text generation metrics on the test set.}
  \centering
  \resizebox{0.55\textwidth}{!}{
    \begin{tabular}{l|lccccc}
      \toprule
      & & \small BLEU@1$\uparrow$ & \small BLEU@4$\uparrow$ & \small ROUGE$\uparrow$ & \small BERTScore$\uparrow$ \\
      \midrule
      \multirow{5}{*}{\rotatebox[origin=c]{90}{\parbox{1.0cm}{\centering\small H3D}}} 
       & TM2T & $48.90$ & $8.27$ & $38.1$ & $32.2$ \\
       & MotionGPT & $48.20$ & $12.47$ & $37.4$ & $32.4$ \\
       & MotionChain & $48.10$ & $\bm{12.56}$ & $33.9$ & $36.9$ \\
       & MotionGPT3 & $51.06$ & $8.43$ & $38.7$ & $32.0$ \\
       & MG-MotionLLM & $-$ & $8.06$ & $-$ & $36.7$ \\
       & MoGIC (ours) & $\bm{53.13}$ & $10.36$ & $\bm{40.6}$ & $\bm{40.7}$ \\
      \midrule
      \multirow{3}{*}{\rotatebox[origin=c]{90}{\parbox{1.0cm}{\centering\small Mo440H}}} 
       & T2MT & $28.99$ & $15.37$ & $36.22$ & $29.01$ \\
       & MG-MotionLLM & $35.47$ & $17.97$ & $39.07$ & $30.95$ \\
       & MoGIC (ours) & $\bm{42.52}$ & $\bm{20.32}$ & $\bm{39.31}$ & $\bm{31.96}$ \\
      \bottomrule
    \end{tabular}
  }
  \label{tab:m2t-all}
  \vspace{-0.5cm}
\end{wraptable}

\subsection{Downstream Applications}

Following cross-modal generative training, MoGIC supports arbitrary multimodal inputs (language, vision, motion) to produce unified outputs in motion sequences and intentions. Further finetuning on specific tasks enhances performance in specialized settings. We evaluate on HumanML3D~\cite{guo2022humanml3d} and our integrated dataset Mo440H. For HumanML3D, we adopt evaluators from prior work~\cite{meng2025mardm}. For the integrated dataset, we train an evaluator on Mo440H following the previous methods~\cite{guo2022humanml3d}.

\paragraph{Motion Generation and Caption}
We evaluate language-to-motion generation both with and without fine-tuning, as well as motion captioning after finetuning. Experiments are conducted on HumanML3D and Mo440H-ML.

For motion generation on HumanML3D, we adopt a single-stage training strategy, jointly optimizing the motion generation loss and the intent prediction loss, achieving substantial improvements over state-of-the-art methods in terms of FID and R-Precision (Tables~\ref{tab:t2m-h3d}). Results on Mo440H are shown in Tables~\ref{tab:t2m-mydataset}, where MoGIC denotes the model trained solely through cross-modal generative learning, and MoGIC \textit{w/ FT} represents the variant further fine-tuned on the language-to-motion task. We also present the results without computing the generation loss conditioned on the visual modality (denoted as MoGIC \textit{T2M + Int. loss} in the Table~\ref{tab:t2m-mydataset}). As shown, language-based motion generation achieves better results, but its functionality remains limited. All evaluations are conducted using our retrained evaluator on the Mo440H dataset, following the same protocol as previous work~\cite{guo2022humanml3d}.

In addition, Table~\ref{tab:m2t-all} reports results for fine-tuning on motion caption task. During fine-tuning, we feed the entire motion sequence as input and generate textual descriptions. Compared with LLM-based methods, the Intent Prediction Head (IPH) in MoGIC is highly lightweight and does not rely on pre-trained language models, yet it still delivers competitive and effective performance.

\paragraph{Motion In-Between}
We evaluate our method on the motion in-between task, which generates plausible transitions from partial motion contexts. We consider prefix, suffix, infix, and circumfix completion, predicting missing segments at the beginning, end, middle, or both ends of a motion sequence. Experiments on HumanML3D and Mo440H, compared with MARDM~\cite{meng2025mardm}, are reported under two settings: (i) in-between with language, using both textual descriptions and visible motion fragments, and (ii) in-between without language, using only motion fragments. Without task-specific fine-tuning, our method consistently outperforms baselines, as shown in Table~\ref{tab:in-between-all}.

\begin{wrapfigure}{r}{0.38\textwidth}
% \vspace{-1.5cm}
    \centering
    \includegraphics[width=0.35\textwidth]{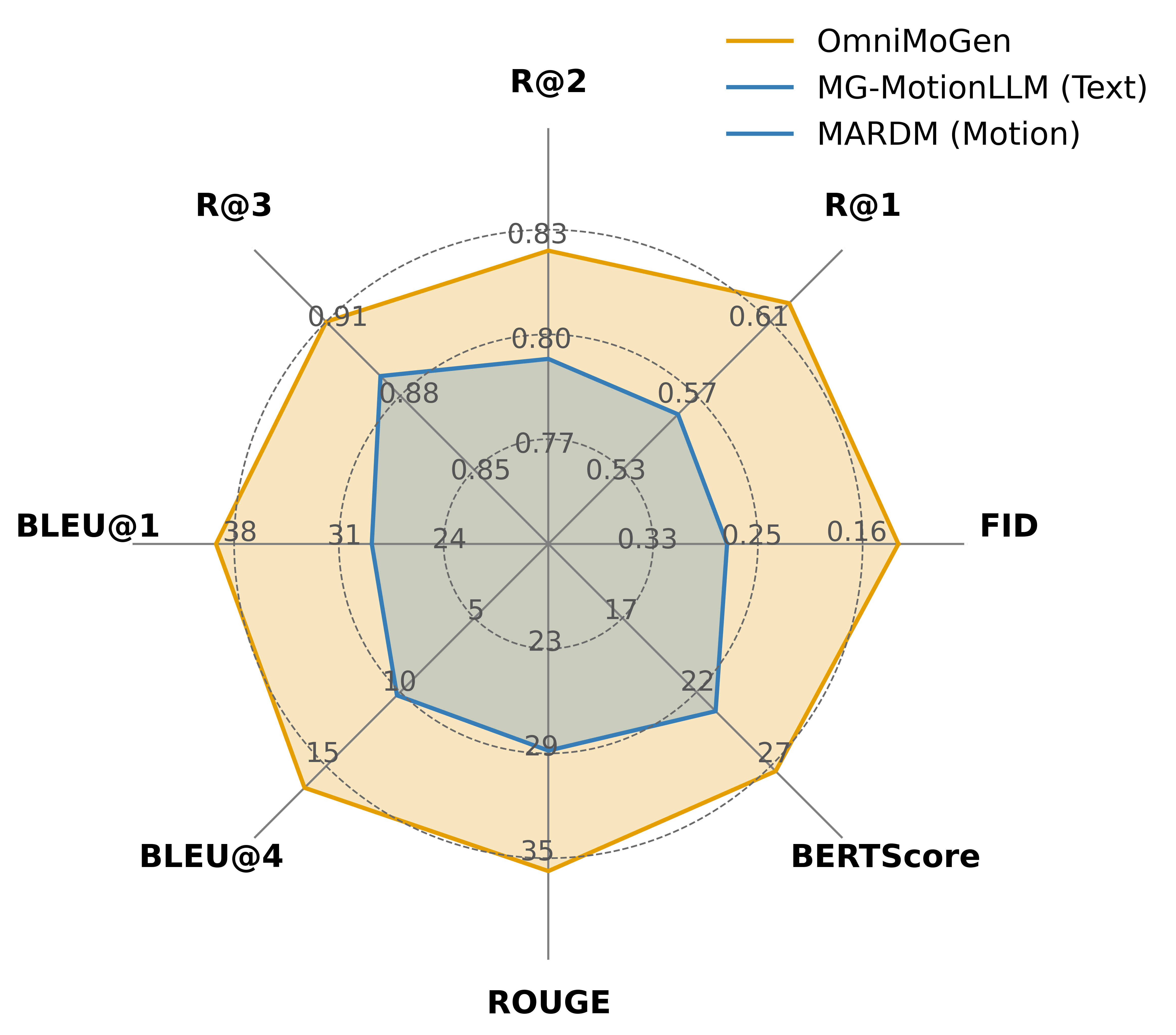} 
    \caption{Comparisons of intention prediction results.}
    \label{fig:intention-prediction}
% \vspace{-0.5cm}
\end{wrapfigure}

\paragraph{Intention Prediction}
The intention prediction task requires the model to infer the conceptual goal of an observed motion and express it in textual form. Given the first 50\% of a motion sequence, the model outputs a complete language description that conveys the underlying intention. Meanwhile, MoGIC can also generate a future motion sequence aligned with this intention. We train two baselines separately for intention prediction~\cite{wu2025mg} and future motion generation~\cite{meng2025mardm}. Without fine-tuning, our model surpasses both, achieving higher quality in intention prediction and lower FID for the synchronously generated motion continuation, as shown in Figure~\ref{fig:intention-prediction}.

\begin{figure}[t]
\centering 
\includegraphics[width=1\columnwidth]{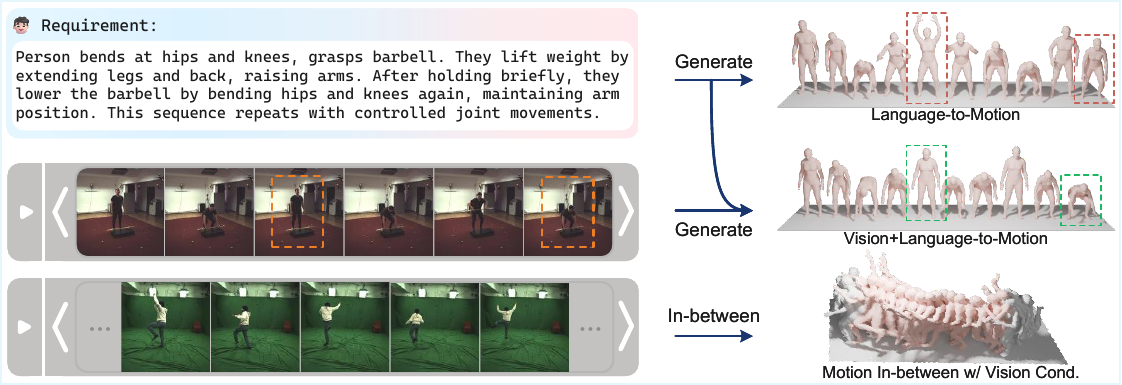} 
\vspace{-0.5cm}
\caption{Visualization of motion generation and motion in-between tasks with vision modality.}
\vspace{-0.5cm}
\label{fig:vision}
\end{figure}

\paragraph{Vision-Augmented Tasks}
We further extend our framework to vision-augmented scenarios, where image sequences serve as additional conditions for motion generation. We focus on two representative tasks: (i) vision-language-to-motion, where textual descriptions and visual frames jointly guide motion synthesis, and (ii) vision-based motion in-between, where visual cues complement partial motion fragments to complete missing segments. These tasks provide a natural and accessible source of conditioning signals that enrich the controllability of generated motions. As shown in Figure~\ref{fig:vision}, when generating a weightlifting motion conditioned only on the text prompt ``lift weight by extending legs and back, raising arms'', the description neither specifies the exact position of the barbell nor provides the model with a prior about the abstract concept of weight. As a result, the model produces an unrealistic sequence in which the barbell is lifted overhead, which is clearly inconsistent with real-world biomechanics. By incorporating visual modality, however, the model gains explicit information about the barbell’s position relative to the body, allowing it to generate natural lifting motions that adhere to realistic constraints.

\begin{table}[t]
\scriptsize
\centering
\caption{The effectiveness of the vision modality. We evaluate MoGIC on the motion generation and in-between tasks conditioning on different conditions without finetuning. \textit{L, V, M} represent language, vision, and motion, respectively.}
\begin{adjustbox}{width=\textwidth}
\begin{tabular}{llcccccc}
\toprule
\multirow{2}{*}{Category} & \multirow{2}{*}{Task} & 
\multicolumn{3}{c}{R Precision$\uparrow$} & 
\multirow{2}{*}{FID$\downarrow$} & 
\multirow{2}{1cm}{\centering Matching$\downarrow$} & 
\multirow{2}{1.5cm}{\centering Diversity$\uparrow$} \\
\cline{3-5}
& & Top 1 & Top 2 & Top 3 \\
\midrule

\multirow{3}{*}{Motion Gen.} 
& \textit{L2M} & $0.590^{\pm 0.002}$ & $0.804^{\pm 0.001}$ & $0.897^{\pm 0.002}$ & $0.330^{\pm 0.003}$ & $1.913^{\pm 0.007}$ & $12.757^{\pm 0.075}$ \\
& \textit{V2M} & $0.408^{\pm 0.002}$ & $0.639^{\pm 0.003}$ & $0.789^{\pm 0.003}$ & $0.634^{\pm 0.010}$ & $2.881^{\pm 0.012}$ & $12.662^{\pm 0.108}$ \\
& \textit{LV2M} & $0.589^{\pm 0.001}$ & $0.801^{\pm 0.001}$ & $0.898^{\pm 0.002}$ & $0.266^{\pm 0.002}$ & $1.953^{\pm 0.007}$ & $12.585^{\pm 0.067}$ \\
\midrule
\multirow{4}{*}{Motion In-Bet.} 
& prefix & $0.498^{\pm 0.002}$ & $0.720^{\pm 0.002}$ & $0.830^{\pm 0.001}$ & $0.436^{\pm 0.004}$ & $2.373^{\pm 0.006}$ & $12.469^{\pm 0.052}$ \\
& prefix \textit{w/ L} & $0.624^{\pm 0.001}$ & $0.833^{\pm 0.001}$ & $0.918^{\pm 0.001}$ & $0.137^{\pm 0.001}$ & $1.707^{\pm 0.003}$ & $12.487^{\pm 0.041}$ \\
& prefix \textit{w/ V} & $0.553^{\pm 0.001}$ & $0.766^{\pm 0.001}$ & $0.868^{\pm 0.001}$ & $0.205^{\pm 0.001}$ & $2.021^{\pm 0.004}$ & $12.639^{\pm 0.058}$ \\
& prefix \textit{w/ L+V} & $0.619^{\pm 0.001}$ & $0.830^{\pm 0.001}$ & $0.914^{\pm 0.001}$ & $0.132^{\pm 0.001}$ & $1.701^{\pm 0.004}$ & $12.662^{\pm 0.065}$ \\
\bottomrule
\end{tabular}
\end{adjustbox}
\vspace{-0.5cm}
\label{tab:ablation-vision}

\end{table}

\subsection{Ablation Study}
\paragraph{Effectiveness of Intention Prediction}
Ablation results on HumanML3D (MoGIC \textit{w/o Int. loss} in Table~\ref{tab:t2m-h3d}) and Mo440H (MoGIC \textit{only T2M loss} in Table~\ref{tab:t2m-mydataset}) show that removing intention prediction consistently lowers performance, with the largest drops in FID ($-35.2\%$ on HumanML3D, $-33.3\%$ on Mo440H) and retrieval precision ($-0.9\%$ on HumanML3D, $-1.8\%$ on Mo440H). We further replace the intention prediction loss with a captioning loss, training the model to generate descriptions from complete motion sequences (MoGIC \textit{L2M + Caption loss}). Caption supervision improves motion quality, but the gains are notably smaller than those from intention prediction. This underscores that intention modeling is crucial for producing high-quality motion. Without it, the model relies on shallow correlations, failing to capture the causal structure and latent goals of human motion. Consequently, generated motions lose realism. Moreover, the improved performance from jointly optimizing intention prediction and motion generation underscores that modeling the understanding of motion is key to generating more faithful results.

\begin{wrapfigure}{r}{0.55\textwidth}
    % \vspace{-1cm}
    \captionsetup{skip=2pt}
    \centering
    \begin{subfigure}{0.93\linewidth}
        \centering
        \includegraphics[width=\linewidth]{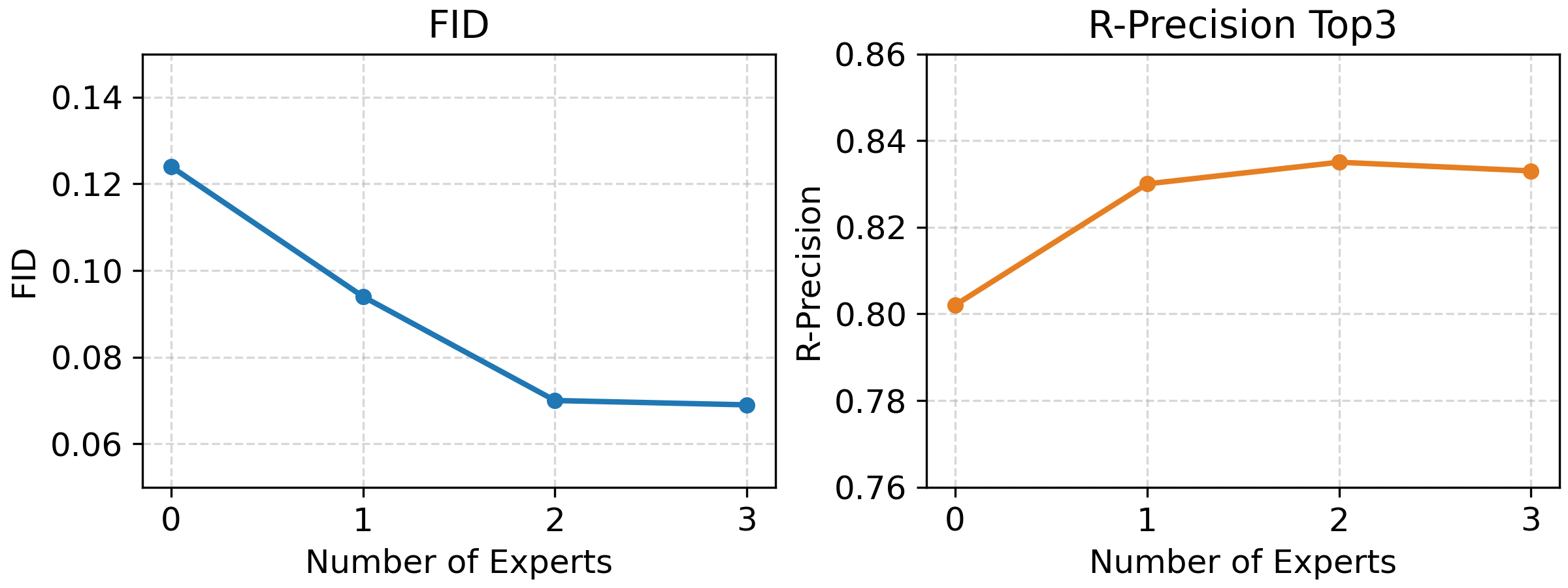}
        \caption{Ablation study of the mixture-of-attentions}
        \label{fig:ablation-attention}
    \end{subfigure}
    \begin{subfigure}{0.92\linewidth}
        \centering
        \includegraphics[width=\linewidth]{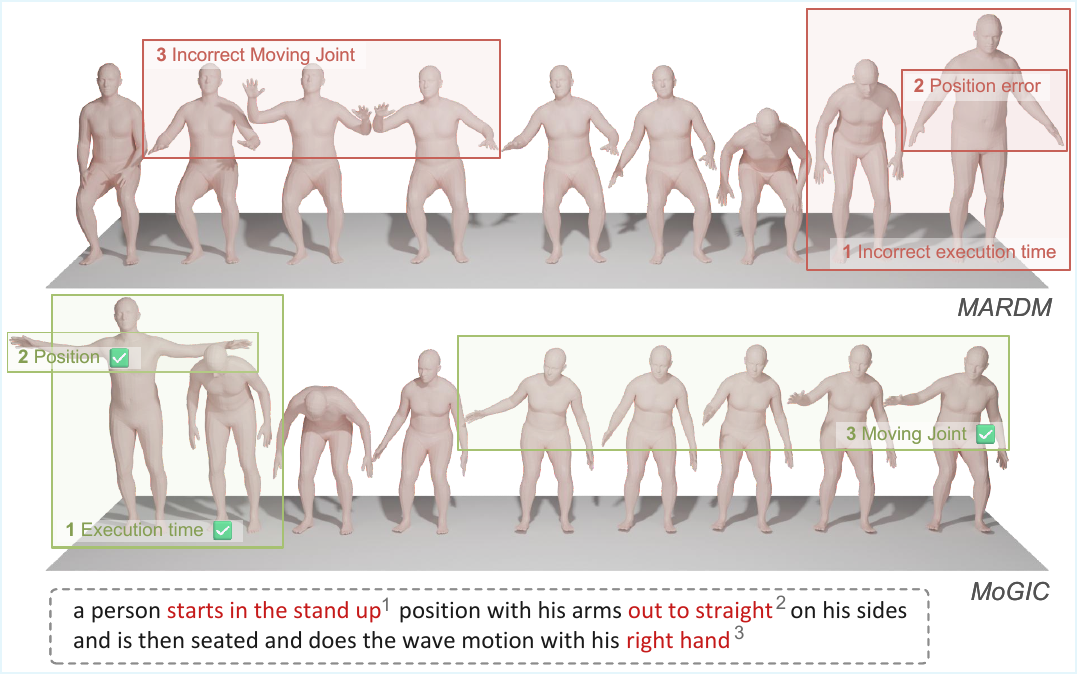}
        \caption{Visualization results of L2M.}
        \label{fig:ablation-visualization}
    \end{subfigure}
    \caption{The effectiveness of mixture-of-attention.}
    \label{fig:ablation}
    % \vspace{-1cm}
\end{wrapfigure}

\paragraph{Effectiveness of Training with Vision Modality}
We assess the contribution of vision from two complementary angles. (i) \textit{Vision as priors.} We drop vision-to-motion and vision–language-to-motion losses in the cross-modal generative training (MoGIC \textit{L2M + Int. loss} in Table~\ref{tab:t2m-mydataset}). Compared to MoGIC \textit{w/ L2M FT} which is finetuned on language-to-motion and intention prediction losses after the complete cross-modal generative training, training without vision modality leads to degraded language-to-motion performance, indicating that the visual modality enables the model to learn richer contextual representations and implicitly guides the alignment between generated motions and their conditioning inputs. (ii) \textit{Vision as a conditioning modality.} We further examine whether adding vision conditions improves generation. On Mo440H-MLV, we evaluate both vision–language-to-motion and vision-based motion in-between without task-specific fine-tuning. As shown in Table~\ref{tab:ablation-vision}, vision consistently reduces FID while keeping diversity comparable, and combining language with vision yields the best trade-off. This suggests that visual conditions provide complementary spatiotemporal hints beyond text or motion alone, leading to more natural and coherent generations.

\paragraph{Effectiveness of Mixture-of-Attention}
e, we test four settings: no expert (i.e., no cross-attention with fine-grained conditions), one expert ($k_1 \in [0, \infty)$), two experts ($k_1 \in [0, \infty), k_2 \in [1, 6)$), and three experts ($k_1 \in [0, \infty), k_2 \in [1, 6), k_3 \in [6, 10)$). As shown in Figure~\ref{fig:ablation-attention}, fine-grained conditions greatly boost retrieval performance. Increasing expert number steadily reduces FID, with retrieval precision peaking at two experts. To balance efficiency and effectiveness, we adopt two experts as default. Figure~\ref{fig:ablation-visualization} further shows that, thanks to mixture-of-attention with adaptive scope, our method generates motions with more precise local responses, including joint movement, positioning, and timing.

% \paragraph{Effectiveness of Mixture-of-Attention}

% \paragraph{Motion Generation (MG)} Text-Conditioned MG, Vision-Conditioned MG, Text-Vision-Conditioned MG
% \paragraph{Motion In-between (MI)} MI, Text-Conditioned MI, Vision-Conditioned MI.
% \paragraph{Joint Intention and Motion Generation (IMG)} Vision-Conditioned IMG, Visible Motion-Conditioned IMG
% \paragraph{Motion Caption}
\section{Conclusion}
In this work, we present MoGIC, a unified framework that integrates intention modeling and visual priors into multimodal motion generation. Through joint training of intention prediction and motion synthesis across language, vision, and motion, MoGIC uncovers latent human goals and achieves versatile generative capability. To support this paradigm,  we curated and automatically annotated Mo440H, a large-scale tri-modal benchmark comprising 440 hours of diverse human motions. Extensive experiments validate the strong performance and broad functionality of MoGIC, demonstrating significant gains in both motion fidelity and text generation tasks. We believe this work provides new insights into multimodal human motion generation and understanding, and lays a foundation for future research toward more precise, adaptive, and intention-aware motion synthesis.

\bibliography{iclr2026_conference}
\bibliographystyle{iclr2026_conference}

\newpage

\appendix

\renewcommand{\thetable}{A\arabic{table}}
\renewcommand{\thefigure}{A\arabic{figure}}
\setcounter{figure}{0}
\setcounter{table}{0}

\section{Appendix}

\begin{figure}[t]
\centering 
\includegraphics[width=1\columnwidth]{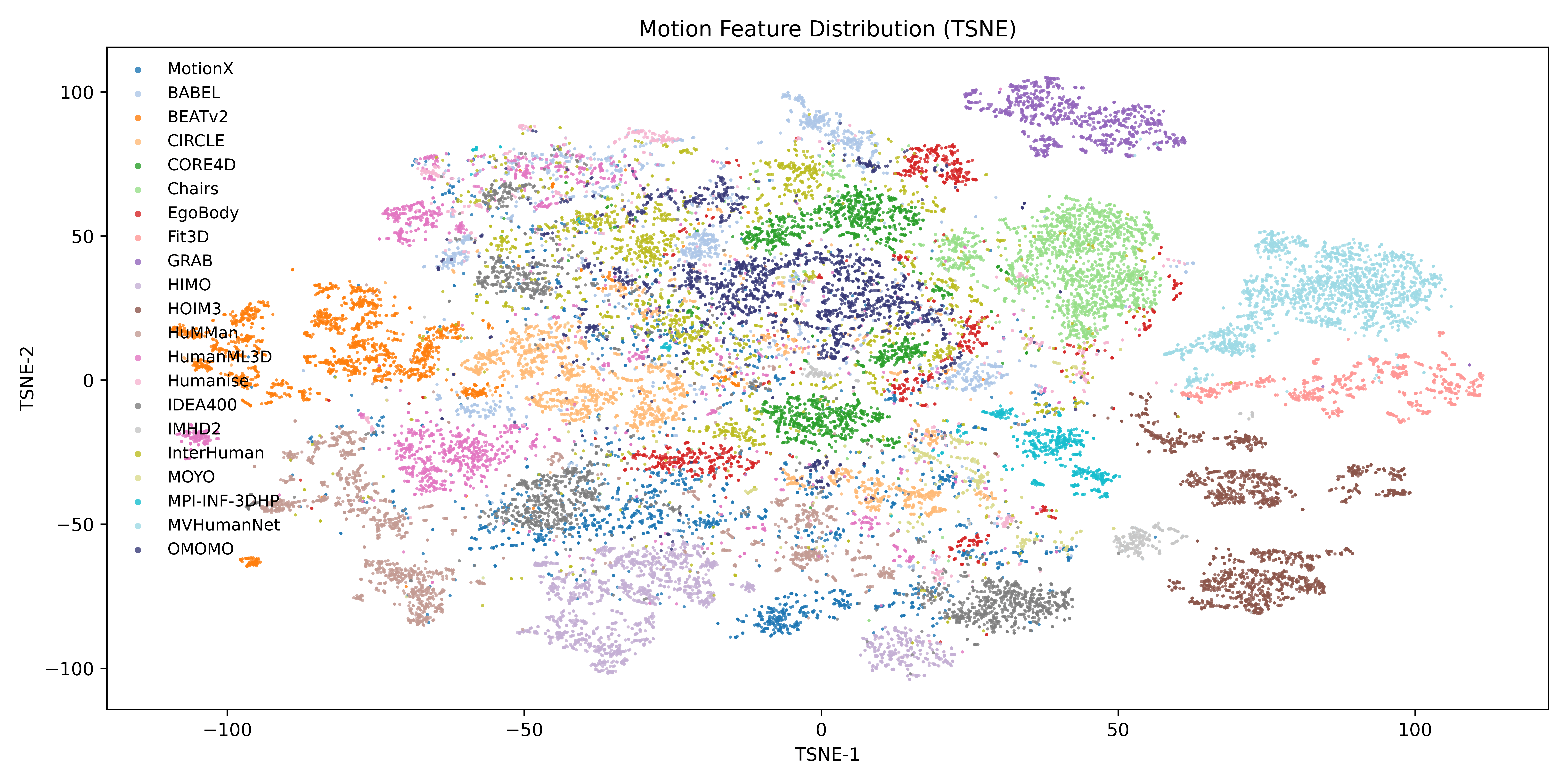} 
\caption{Visualization of data distributions. For each dataset, we randomly sample 2,000 motion sequences. Each sequence is temporally averaged to obtain a compact feature representation, which is then reduced in dimensionality using t-SNE.}
\label{fig:dataset_distribution}
\end{figure}

\subsection{Mo440H Motion Dataset}
Some existing approaches estimate human motion directly from web videos. However, such video data often suffer from motion degradation caused by low image quality, occlusions, and frequent viewpoint changes, which result in unstable monocular pose estimation and ultimately compromise the quality of motion generation at the data level. In addition, these datasets typically exhibit limited motion diversity. For instance, many MotionX~\citep{lin2023motionx} samples come from repetitive video series (e.g., yoga tutorials), while MotionMillion~\citep{fan2025go} is dominated by sports-related activities such as martial arts, fitness, and dance.

To ensure both data quality and coverage across diverse motion scenarios, we construct Mo440H, a unified benchmark built from 21 high-quality open-source motion datasets. We standardize properties such as frame rate, motion orientation, and maximum sequence length, while integrating motions spanning single-person activities, human–object interactions, and multi-person interactions. The included datasets are as follows. Furthermore, we visualize the dataset distribution in Figure~\ref{fig:dataset_distribution}. For each dataset, we randomly sample 2,000 motion sequences, temporally average each sequence into a compact feature representation, and then apply t-SNE for dimensionality reduction.

\textit{BABEL~\citep{punnakkal2021babel}} A large-scale motion-language dataset (~43 h) from AMASS, providing 28k labeled sequences and 63k frame-level annotations across 250 categories for action recognition, segmentation, and motion-language understanding.

\textit{BEAT (v2)~\citep{liu2024emage}} A 76 h multi-modal co-speech gesture dataset with 30 speakers in 4 languages, offering synchronized body motion, facial expression, audio, text, and emotion for gesture generation research.

\textit{Fit3D (AIFit)~\citep{fieraru2021aifit}} A fitness motion dataset with 13 subjects performing 37 exercises, containing 3 M RGB frames aligned with 3D poses, supporting pose estimation and automated training feedback.

\textit{HuMMan~\citep{cai2022humman}} A massive multi-modal 4D dataset of 1000 subjects and 500 actions (~60 M frames), providing synchronized RGB, depth, IMU, skeletons, SMPL, and textured meshes for comprehensive human modeling.

\textit{HumanML3D~\citep{guo2022humanml3d}} A motion-language dataset with 14,616 motion clips paired with 44,970 captions, enabling text-conditioned motion generation and motion-to-language tasks.

\textit{IDEA-400~\citep{lin2023motionx}} A dataset of ~13k sequences across 400 action categories with text descriptions, offering a benchmark for generalizable motion generation and recognition.

\textit{MoYo~\citep{tripathi20233d}} A yoga-focused dataset with ~200 poses (1.75 M frames) captured by multi-camera and pressure-sensing mat, providing SMPL-X fits and contact data for studying balance and extreme poses.

\textit{MPI-INF-3DHP~\citep{mehta2017monocular}} A 3D pose dataset of diverse single-person activities in indoor/outdoor settings, captured markerlessly with multi-cameras for in-the-wild pose estimation benchmarks.

\textit{MVHumanNet~\citep{xiong2024mvhumannet}} A large-scale multi-view dataset with 4,500 identities and 60k sequences (645 M images), annotated with segmentation and calibration for digital human reconstruction under clothing variations.

\textit{Motion-X~\citep{lin2023motionx}} A 3D whole-body dataset of ~96k sequences (13.7 M frames) in SMPL-X format, capturing expressive motions with hand and face details, with text labels for >80k sequences. Although this dataset is derived from web data, we still include it in Mo440H to enhance diversity.

\textit{EgoBody~\citep{zhang2022egobody}} An egocentric multi-person dataset with 125 sequences, combining HoloLens2 RGB-D, gaze, and ground-truth 3D meshes for first-person interaction understanding.

\textit{InterHuman (InterGen)~\citep{liang2024intergen}} A large-scale two-person interaction dataset (~107 M frames) with 16,756 textual descriptions, covering diverse social and cooperative motions.

\textit{HOI-M$^3$~\citep{zhang2024hoi}} A multi-person multi-object dataset (~181 M frames) of 46 subjects and 90 objects, recorded by 42 cameras and IMUs for studying complex group interactions.

\textit{CORE4D~\citep{liu2024core4d}} A dataset of 1,000 real-world and 10k+ augmented collaborative rearrangement sequences, focusing on multi-human multi-object cooperation in household scenes.

\textit{CIRCLE~\citep{araujo2023circle}} A contextual dataset of 10 h reaching/interaction motions across 9 scenes with synchronized egocentric RGB-D, enabling study of human–scene relations.

\textit{CHAIRS~\citep{jiang2023full}} A dataset of 17.3 h interactions between 46 people and 81 articulated chairs, with aligned human–object meshes for studying posture and manipulation.

\textit{GRAB~\citep{taheri2020grab}} A whole-body grasp dataset with 10 subjects interacting with 51 objects, providing detailed body, hand, face, and object contact data for realistic manipulation modeling.

\textit{HIMO~\citep{lv2025himo}} A benchmark with 3,300 sequences (4.08 M frames) of long, multi-object interactions paired with fine-grained text annotations for complex activity understanding.

\textit{HUMANISE~\citep{wang2022humanise}} A synthetic dataset aligning mocap motions with 3D indoor scenes, annotated with language descriptions to enable motion generation conditioned on scene context.

\textit{IMHD$^2$~\citep{zhao2024imhoi}} A high-dynamic interaction dataset (295 sequences, ~892k frames) captured with high-speed cameras and IMUs, featuring fast-motion tasks with ground-truth body and object trajectories.

\textit{OMOMO~\citep{li2023object}} A conditional dataset of ~10 h sequences coupling object trajectories with human reactions, providing human–object pairs for learning motion conditioned on object dynamics.

\begin{figure}[t]
\centering 
\includegraphics[width=1\columnwidth]{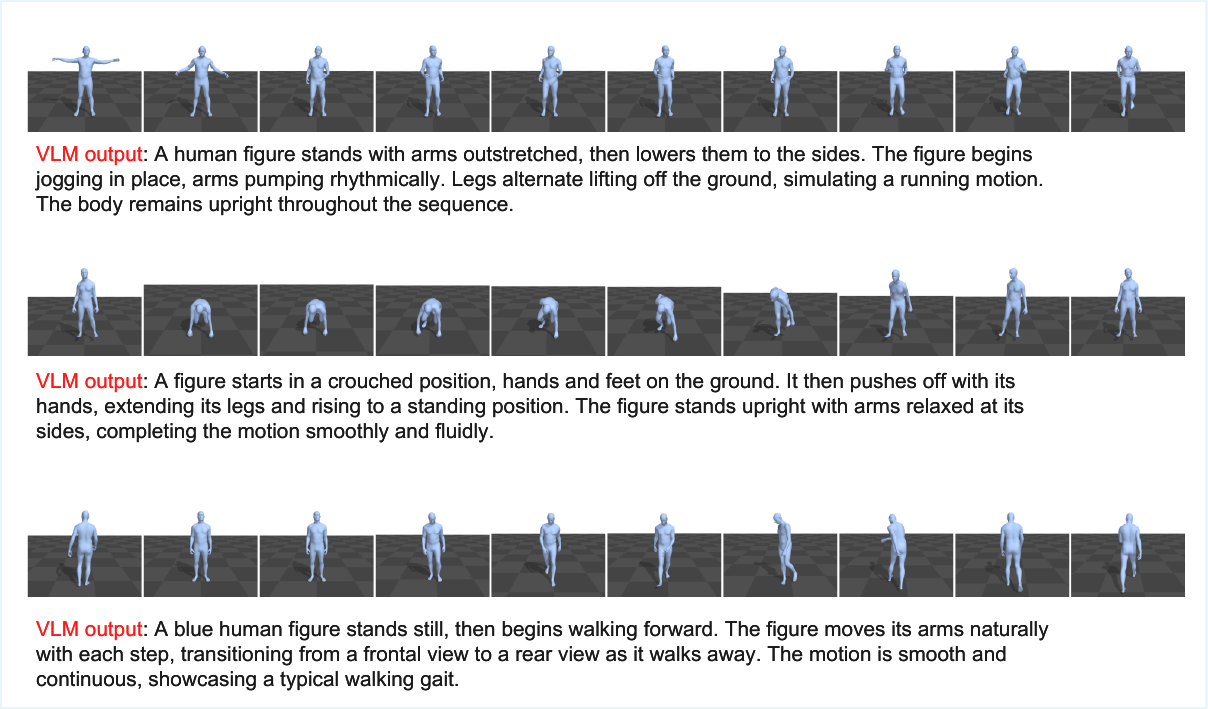} 
\caption{Some examples of motion descriptions generated by Qwen2.5-VL-Max~\citep{ali2025Qwen2.5-VL}. Although the input is a human mesh sequence rendered based on SMPL parameters, the Qwen model still outputs a relatively accurate description.}
\label{fig:qwen_caption}
\end{figure}

For datasets that contain visual modality but lack textual annotations, we first segment motion sequences with a maximum length of 10 seconds and then use Qwen2.5-VL-Max to automatically generate descriptions for each sequence. The annotation prompt is as follows:

\vspace{1em}
\textit{\textcolor{gray}{Briefly describe the human motion, focusing on the interaction with objects and body movements. Do not describe text that appears in the video. Describe objectively. The output must be in one paragraph and no more than 100 words. Do not describe the time explicitly.}}
\vspace{1em}

Some datasets do not provide RGB video modality; instead, they render virtual motion videos from SMPL body representations. For these datasets, we also employ Qwen to perform annotations. We observe that Qwen demonstrates strong robustness in handling such virtual humans and produces reasonably accurate motion descriptions. Figure~\ref{fig:qwen_caption} illustrates several annotation examples on rendered virtual humans.

\subsection{Evaluation Metrics}
For motion generation, we use several common quantitative metrics to evaluate different aspects of model performance. FID (Fréchet Inception Distance) measures how close the distribution of generated motions is to real motions, reflecting overall realism. R-Precision and Matching Score evaluate whether the generated motions match the given text descriptions, focusing on text–motion alignment. To check the diversity of outputs, we report Multimodality Diversity, which measures how much variation the model can produce under the same condition. We also include CLIP-Score (following~\cite{meng2025mardm}), which leverages pretrained vision–language models to further test semantic consistency between motion and text.

For text generation, we evaluate the captions generated for motions using both n-gram and semantic metrics. BLEU@4 and BLEU@1 measure overlap at different n-gram levels, while ROUGE emphasizes recall of important phrases. To capture similarity beyond exact wording, we report BERTScore, which uses contextual embeddings to measure semantic closeness to reference captions.

\begin{figure}[t]
\centering 
\includegraphics[width=1\columnwidth]{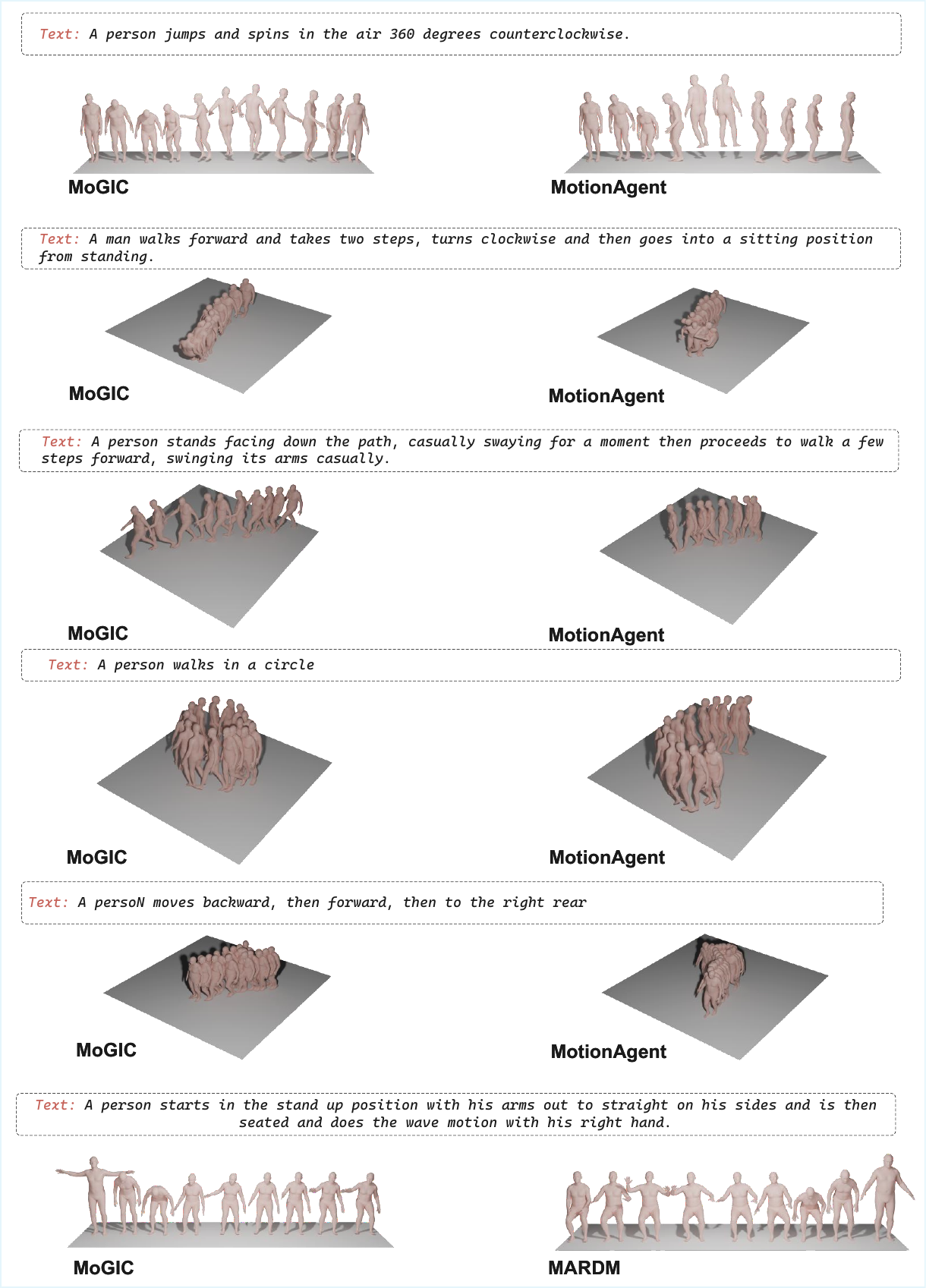} 
\caption{Visualization results of language-to-motion generation.}
\label{fig:vis_h3d}
\end{figure}

\begin{figure}[t]
\centering 
\includegraphics[width=1\columnwidth]{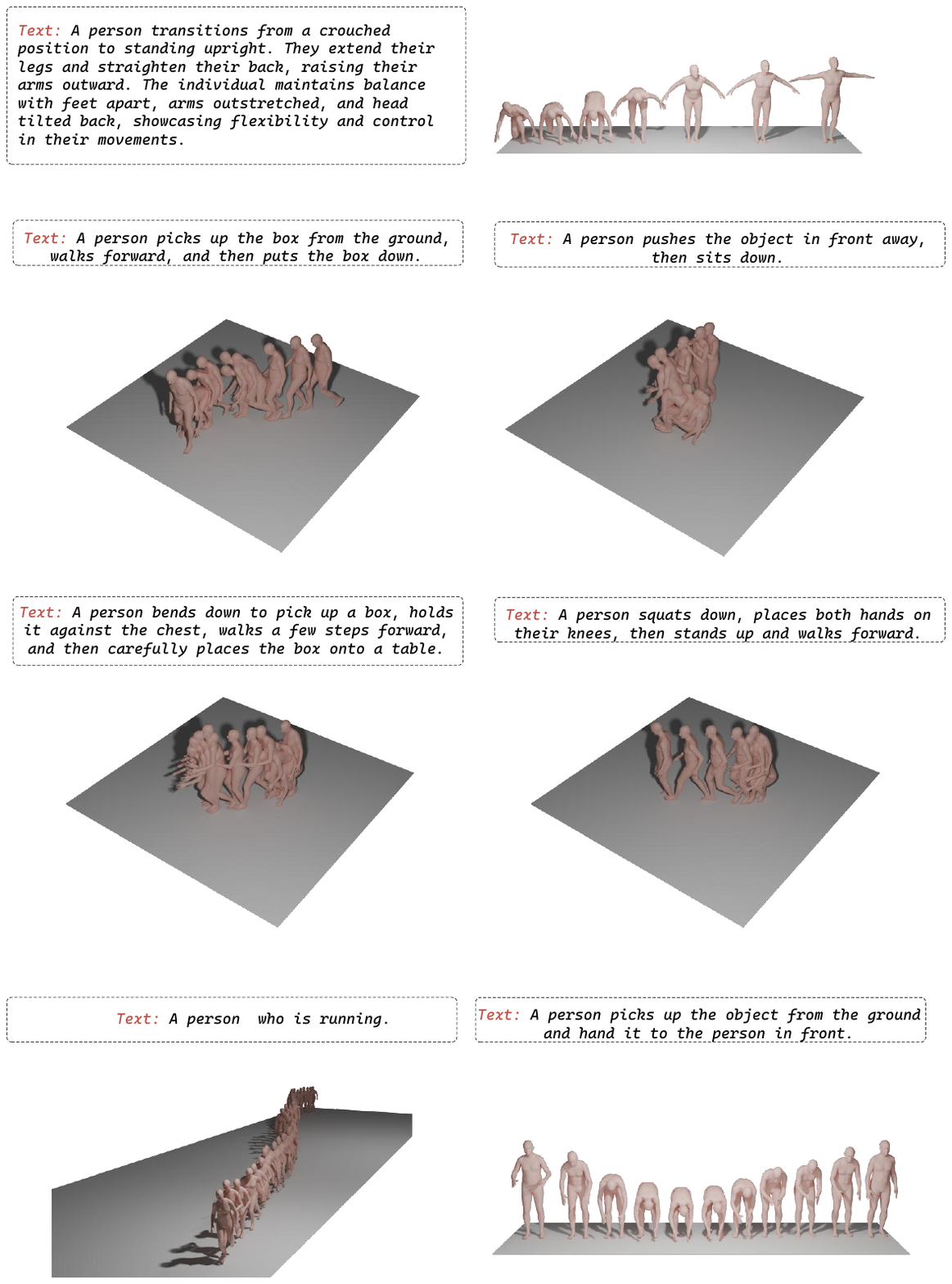} 
\caption{Visualization results of language-to-motion generation.}
\label{fig:vis_mo440h}
\end{figure}

\subsection{Visualization Results}
Figure~\ref{fig:vis_h3d} presents a comparison of language-to-motion training and testing results on the HumanML3D dataset. We visualize MoGIC alongside MARDM and MotionAgent. As shown, motions generated by MoGIC exhibit more coherent behavioral logic and higher realism. Figure~\ref{fig:vis_mo440h} further provides zero-shot inference examples after training on Mo440H, where MoGIC performs particularly well on tasks with causal dependencies, such as picking up and placing objects.

\begin{table}[t]
\tiny
\centering
\caption{Ablation study of iteration number in masked modeling and sampling steps in diffusion.}
\begin{adjustbox}{width=\textwidth}
\begin{tabular}{lcccccccc}
\toprule
\multirow{2}{*}{Method} & \multirow{2}{*}{Sampling Steps} & 
\multirow{2}{1.5cm}{\centering Iter. Number} & 
\multicolumn{3}{c}{R Precision$\uparrow$} & 
\multirow{2}{*}{FID$\downarrow$} & 
\multirow{2}{1cm}{\centering Matching$\downarrow$} & 
\multirow{2}{1cm}{\centering Inference Time(s)} \\
\cline{4-6}
& & & Top 1 & Top 2 & Top 3 \\
\midrule

\multirow{8}{*}{MoGIC} 
& $20$ & $17$ & $0.540$ & $0.740$ & $0.832$ & $0.079$ & $3.02$ & $1.883$ \\
& $10$ & $17$ & $0.545$ & $0.741$ & $0.835$ & $0.070$ & $3.00$ & $0.840$ \\
& $8$ & $17$ & $0.530$ & $0.735$ & $0.829$ & $0.071$ & $3.02$ & $0.683$ \\
& $5$ & $17$ & $0.525$ & $0.733$ & $0.826$ & $0.068$ & $3.04$ & $0.418$ \\
& $2$ & $17$ & $0.509$ & $0.705$ & $0.813$ & $0.378$ & $3.15$ & $0.193$ \\
& $10$ & $5$ & $0.545$ & $0.742$ & $0.840$ & $0.138$ & $2.98$ & $0.255$ \\
& $10$ & $10$ & $0.534$ & $0.742$ & $0.835$ & $0.079$ & $3.01$ & $0.493$ \\
& $10$ & $25$ & $0.537$ & $0.734$ & $0.837$ & $0.073$ & $3.01$ & $1.095$ \\
\midrule
\multirow{1}{*}{MARDM} 
& - & $17$ & $0.500$ & $0.695$ & $0.795$ & $0.114$ & $3.270$ & $9.813$ \\
\midrule
\multirow{1}{*}{MMM} 
& - & $10$ & $0.487$ & $0.683$ & $0.782$ & $0.132$ & $3.359$ & $0.358$ \\
\bottomrule
\end{tabular}
\end{adjustbox}
\label{tab:ablation-inference}
\end{table}

\subsection{Inference Speed}
All inference experiments are conducted on a workstation equipped with an AMD 9950X CPU and an NVIDIA RTX 5090 GPU. We measure inference efficiency as the wall-clock latency for generating 32 samples in parallel. We analyze two major factors that affect inference speed and performance:

\paragraph{Number of Iterations.}
Our model consists of a conditional masked transformer and a diffusion head and the transformer is followed the training pipeline of masked modeling~\citep{meng2025mardm, pinyoanuntapong2024mmm}. Inference starts from a fully masked latent variable, and the model reconstructs the masked regions iteratively. We set the default number of iterations to 17, and further vary this value to examine its effect on inference speed and performance. Results are reported in Table~\ref{tab:ablation-inference}.

\paragraph{Number of Diffusion Sampling Steps.}
The diffusion module adopts Euler sampling with 10 default steps. Unlike MARDM~\citep{meng2025mardm}, which relies on adaptive solvers (dopri5), our approach generates realistic motions with only a small number of steps. We also compare different numbers of sampling steps and report the corresponding runtime and performance in Table~\ref{tab:ablation-inference}.

These experiments demonstrate that, compared with other diffusion-based methods, our model achieves a clear advantage in real-time inference, maintaining competitive performance even with very few sampling steps.

\section*{LLM Usage Statement}

We made limited use of large language models (LLMs) during the preparation of this work, and we accept full responsibility for all content presented in the paper.

\paragraph{Writing support.} 
LLMs (Qwen2.5 and Qwen3) were used to suggest alternative phrasings and to refine grammar in early drafts of the abstract, introduction, and related work sections. All technical descriptions, theoretical claims, and citations were written and verified solely by the authors.

\paragraph{Research support.} 
LLMs were occasionally employed for ideation, such as exploring possible ablation settings and generating generic utility code (e.g., configuration scripts and logging utilities). All experimental code, implementations of the proposed method, and reported results were developed, validated, and confirmed by the authors.

\paragraph{Data and annotation.} 
For automated text annotations, we employed the LLM Qwen2.5-VL. The details of this process are described in the paper, and the exact prompts and cached responses are included in the anonymous supplementary materials. Outputs were manually reviewed and corrected where necessary before use.

No LLM has been credited as an author. All outputs were checked for factual accuracy, correctness, and relevance, and the authors remain fully accountable for the entirety of this work.

\end{document}